\documentclass[11pt]{article}

\usepackage{acl}
\usepackage{times}
\usepackage{latexsym}
\usepackage[T1]{fontenc}
\usepackage[utf8]{inputenc}
\usepackage{microtype}
\usepackage{inconsolata}
\usepackage{graphicx}
\usepackage{booktabs}
\usepackage{multirow}
\usepackage{amsmath}
\usepackage{siunitx}
\usepackage{mathtools}
\usepackage[table]{xcolor}
\usepackage{dcolumn}

\newcommand{\sig}{\ensuremath{\downarrow}}
\newcolumntype{d}[1]{D{.}{.}{#1}}
\title{Evaluating Document-Tuned Transformer Representations for Person-level Mental Health Assessment}

\author{Aaron Marker \\
  Vanderbilt University \\
  \texttt{aaron.marker@vanderbilt.edu} \\\And
  Oscar Kjell \\
  Lund University \\\AND
  Vasudha Varadarajan \\
  Carnegie Mellon \\
  University \\\And
  H. Andrew Schwartz \\
  Vanderbilt University \\
  \texttt{hansen.schwartz@vanderbilt.edu} \\}

\begin{document}
\maketitle

\begin{abstract}
Person-level psychological assessment requires aggregating meaning across many messages from the same individual, a task that document-level training objectives were not explicitly designed for. 
We present a systematic, empirical comparison between architecturally matched traditional (a) \textit{base-transformers} and (b) \textit{document-tuned-transformers} (further contrastively fine-tuned at the document-level, sometimes referred to as "sentence transformers") under otherwise identical conditions. 
Comparing layer-wise and overall performance across two longitudinal mental health and psychological datasets,  
we find document-tuned models demonstrated a consistent improvement over base representations (increase in Pearson $r$ of $13.4\%, p=.015$). 
Robustness analyses revealed document-tuned models remained more accurate under perturbations to word deletion, synonym replacement, typo injection, and back translation. 
Further, hedged language (e.g., `usually') was more characteristic of outcomes in document-tuned embeddings while abundance (e.g., `lot') was more characteristic of base-transformers, suggesting document-tuned models may better capture uncertainty.
These results suggest representation choice impacts mental health prediction, document-tuned models often being more adept. 
\end{abstract}

\section{Introduction}
Transformer-based language models are increasingly used to infer psychological constructs such as depression, anxiety, well-being, and personality from text~\cite{tsakalidis2022overview, zirikly2019clpsych, guo2024large, pourkeyvan2024harnessing, zhang2022natural}. Much of prior work relies on token-level pretrained encoders (e.g., BERT, RoBERTa)~\cite{liu2019roberta, devlin2019bert}. 
Recently, sentence transformers trained with document-level contrastive objectives (e.g., Sentence-BERT) have gained popularity for producing semantically meaningful sentence- and document-level embeddings~\cite{reimers2019sentence}.
These embeddings are typically frozen and used in a simple prediction head such that the quality of the representation itself becomes the primary driver of downstream performance.

Document-tuned transformers could bring two important advantages over the base transformers. First, psychological assessments are person-level outcomes~\cite{zhang2022natural, montejo2024survey, garg2023mental} while base transformers ultimately represent information at a much lower level of aggregation, the token. 
Document-level representations already move to a higher level of aggregation (one per document or sentence) where by meaning at the level of whole sequences isn't relegated to words in context. 
Further, representations from traditional word-level transformers are also known to suffer from anisotropy~\cite{ethayarajh-2019-contextual}, a geometric property where embeddings cluster in a narrow cone, rather than being uniformly distributed, often causing unrelated words to appear mathematically similar. 
Document-tuned transformers partially address both of these issues: tuned to produce document-level representations, they aggregate, and bring symmetry across dimensions of embeddings~\cite{gao-etal-2021-simcse, jung2023isotropic}. However, empirically, it remains unclear which representation strategy is better suited for mental health assessment.

The psychological constructs targeted in this study span a broad range of clinical phenomena that differ in how they are experienced and expressed. Internalizing conditions such as depression and anxiety involves affective (e.g., nervous, sad), cognitive (e.g., useless, incapable), and somatic dimensions (e.g., headache, nausea) that individuals may disclose directly \cite{gu2025natural} or express through indirect, hedged language \cite{varadarajan2025linking}. Substance use patterns, are often discussed obliquely — for example, problematic drinking is associated with sadness, frustration, or even describing social events \cite{nilsson2024language}. Understanding how representation strategies may vary in their capacity to capture such diverse signals is essential for psychological prediction — particularly when prediction relies on aggregating many short, naturalistic text responses into a single person-level representation.

We conduct a controlled comparison of MLM-pretrained (base) and contrastively fine-tuned document-level (document-tuned) encoders for user-level psychological prediction. We aim to answer (RQ1) which model is more accurate for psychological prediction? (RQ2) what are optimal layer selection and message aggregation strategies? and (RQ3) do observed differences reflect sensitivity to specific linguistic features or perturbations? Our contributions are: (1) to our knowledge, the first controlled comparison of MLM-pretrained and document-tuned encoders for user-level psychological prediction; (2) an empirical analysis of layer selection and aggregation strategies across these encoder types; and (3) linguistic perturbation experiments that probe the source of observed performance differences in psychological settings.

\section{Related Work}

Much of prior work on transformer-based psychological prediction relies on token-level fine-tuned encoders. Alternative methods explicitly learn document-level representations optimized for semantic similarity. This distinction motivates a focused comparison of how representation choice impacts psychology-oriented prediction tasks.

\subsection{Token level transformer fine-tuning}
Pretrained transformer encoders, such as BERT, established a fine-tuning paradigm in which contextualized token-level models are fine-tuned end-to-end for downstream tasks~\cite{devlin2019bert}. Subsequent models, including RoBERTa, showed that architectural refinements and improved pretraining strategies can substantially improve performance, reinforcing token-level fine-tuning as a strong default across NLP tasks~\cite{liu2019roberta}. For sentence- or document-level prediction, these models typically rely on pooled token embeddings or a special classification token (e.g., [CLS]). However, because pretraining objectives are primarily token-level, these representations are not explicitly optimized for holistic document semantics.

\subsection{Document level representation learning}
A recent development focuses on learning general-purpose document embeddings. Sentence-BERT uses Siamese and contrastive objectives to produce fixed-size representations that capture semantic similarity~\cite{reimers2019sentence}. Unlike token-level fine-tuning, these models directly encode entire sentences or short documents, enabling efficient reuse with lightweight downstream predictors.

\subsection{Language modeling for psychological prediction}
Transformer models have been widely applied in computational psychology, including tasks such as depression detection, suicide risk assessment, and mood prediction, often through shared tasks such as CLPsych ~\cite{tsakalidis2022overview, tseriotou2025overview, zhang2022natural, montejo2024survey, garg2023mental}. While most work relies on token-level pretrained models, document-level encoders (e.g., Sentence-BERT) have also been used, for instance in ensemble systems and shared-task submissions~\cite{ogunleye2024sentiment, azim2022detecting}. The coexistence of these approaches highlights the need for controlled comparisons. We address this by comparing "roberta-large" and "all-roberta-large-v1" under matched extraction and evaluation settings.

\section{Datasets}

We evaluate our models on two longitudinal datasets that differ in scale and outcome focus: (1) a densely sampled ecological momentary assessment dataset (DS4UD) with broad psychological measures~\cite{nilsson2024language}, and (2) a larger longitudinal mental health dataset (LMHD) with a broader user population (Kjell et al., in progress). Both datasets require aggregating multiple temporally distributed text responses to model person-level psychological outcomes, making them well-suited to examine how representation strategies handle aggregation across temporally distributed text.

\begin{table}[t]
\centering
\small
\setlength{\tabcolsep}{6pt}
\renewcommand{\arraystretch}{1.15}

\begin{tabular}{llr}
\toprule
\textbf{Dataset} & \textbf{Statistic} & \\
\midrule


\multirow{3}{*}{DS4UD}
& Total documents (EMAs)       & 10,108 \\
& Average documents per person   & 84.9       \\
& Total people             & 120     \\

\midrule

\multirow{3}{*}{LMHD}
& Total documents (EMAs)     & 7{,}207   \\
& Average documents per person   & 5.5      \\
& Total people              & 1{,}307 \\


\bottomrule

\end{tabular}
\caption{Descriptive statistics for the DS4UD and LMHD datasets under individual-message and concatenated-message representations.}
\label{tab:dataset_stats}
\end{table}

\subsection{The DS4UD Dataset}

DS4UD is a longitudinal EMA dataset of U.S. service industry workers collected via a smartphone application~\cite{nilsson2024language}. Participants (N=120; 75\% female; $M_{age}=35$) completed up to three EMAs per day across multiple 14-day waves. Each EMA included an open-ended English text response (minimum 200 characters) describing the participant’s current affective state, along with self-report ratings of momentary affect (positive and negative affect), perceived stress, recent alcohol use, alcohol craving, and energy.

At the start of each wave, participants completed standardized questionnaires assessing affect such as valence and arousal \cite{remington2000reexamining}, positive and negative affect \cite{thompson2007development}, depressive symptoms (Patient Health Questionnaire–9; PHQ-9) \cite{kroenke2001phq}, anxiety symptoms (Generalized Anxiety Disorder–7; GAD-7) \cite{spitzer2006brief}, perceived stress (Perceived Stress Scale; PSS) \cite{cohen1983global}, alcohol use severity (Alcohol Use Disorders Identification Test; AUDIT, including the AUDIT-C) \cite{saunders1993development}, cravings, exposure to adverse childhood experiences (MACE) \cite{teicher2015maltreatment}, and personality traits using a Big Five inventory (extraversion, agreeableness, conscientiousness, neuroticism, and openness) \cite{soto2017next}. These measures capture clinically relevant internalizing symptoms, substance use risk, stress, and stable personality characteristics. We retain participants who completed at least two waves, yielding over 10{,}000 text-based EMAs from 120 participants.

\subsection{The LMH Dataset}

The LMH Dataset comprise 10-week longitudinal data from a study evaluating alternative approaches to mental health assessment, including measures of general mental health and symptoms related to major depressive disorder (MDD) and generalized anxiety disorder (GAD). Participants were recruited via Prolific, and the present analyses include English-speaking participants only (N = 1,307; M age = 43.9 years, SD = 17.5; male = 34.3\%, female = 64.0\%, other = 1.7\%). The sample was enriched: approximately 50\% of participants were recruited from Prolific’s general population, while the remaining participants were prescreened to report a prior diagnosis of MDD ($\approx 25\%$) or GAD ($\approx 25\%$).

Participants completed comprehensive surveys at baseline and follow-up, with shorter bi-weekly surveys administered in between. Across waves, the study assessed well-being; depressive and anxiety symptoms; perceived stress; alcohol and drug use; trauma-related symptoms; functional impairment; healthcare utilization; and sociodemographics using a combination of brief open-ended paragraph responses, word-selection formats, and standardized rating scales. Validated measures included the PHQ-9 and PHQ-2 for depressive symptoms; the GAD-7 and GAD-2 for anxiety ; subscales from the Inventory of Depression and Anxiety Symptoms (IDAS) assessing dysphoria, suicidality, panic, social anxiety, ill temper, lassitude, insomnia, appetite changes, traumatic intrusions, and well-being \cite{watson2007development}; the Perceived Stress Scale (PSS); the Alcohol Use Disorders Identification Test (AUDIT); the harmony in life score (HILS) \cite{kjell2016harmony}; the Drug Use Disorders Identification Test (DUDIT) \cite{berman2004evaluation}; the Satisfaction with Life Scale (SWLS) \cite{diener1985satisfaction}; and the World Health Organization Disability Assessment Schedule (WHODAS) \cite{ustun2010measuring}. Additional items assessed sick days and mental health service utilization.

In the current study, we use the open-ended response to the general mental health prompt: “How is your mental health? Please describe how you have been over the last two weeks. You can, for example, write about your emotions, thoughts, behaviours, and/or symptoms related to your health. Write at least one paragraph.”

\section{Methods}
We compare representations from (i) a masked language modeling (MLM) pretrained encoder (base) and (ii) a document-level contrastively tuned encoder (document-tuned) built on the same backbone. Representations are evaluated layer-wise under identical extraction, aggregation, and downstream prediction procedures across two longitudinal datasets.

\subsection{Models}
We compare \texttt{roberta-large}~\cite{liu2019roberta}, pretrained with masked language modeling as the base model, and \texttt{all-roberta-large-v1} variant as the document-tuned model, which fine-tunes the same backbone using just over one billion sentence-pairs for contrastive fine-tuning to produce fixed-size embeddings~\cite{reimers2019sentence, allroberta_large_v1_huggingface}.

Both models share architecture and parameter count, isolating representation learning objective as the primary variable. We note that this comparison is not fully controlled for training data, as the document-tuned model was exposed to additional data beyond masked language modeling pretraining. We use publicly available HuggingFace implementations.

\subsection{Representation extraction}
We embed each message individually and mean-pool all message embeddings for a given user to obtain a fixed-length person-level representation.
To assess whether document-tuned encoders benefit from longer context, we also evaluate a concatenation strategy in which all user messages are joined prior to embedding. When concatenated text exceeds the model context window, segments are embedded separately and mean-pooled.

\subsection{Layer-wise representation selection}
We extract and pool representations from every transformer layer of both encoders and train identical downstream models using nested cross-validation (described below). 
Layer-wise evaluation is motivated by prior work showing that linguistic information varies across depth~\cite{tenney2019bert, jawahar2019does, rogers2020primer, liu2019linguistic}. Because document-tuned transformers are explicitly fine-tuned on a semantic similarity objective, they may contain more useful semantic signal in later layers relative to MLM-pretrained models. We therefore compare average Pearson's r score from the 10-fold cross validation, averaging across all outcomes in that dataset, across all layers and select the best-performing layer for each encoder.

\subsection{Predictive modeling and cross-validation}
For each outcome, we fit ridge regression models with nested 10-fold cross-validation, with the regularization parameter $\alpha$ selected via nested cross validation following prior work~\cite{singh2025systematic}. Performance is measured using Pearson’s $r$ between predicted and true values, averaged across folds. The same downstream model and hyperparameter grid are used for all encoder–layer combinations.


\section{Results and Discussion}

\subsection{Overall Accuracy and Optimal layers}
Document-tuned representations outperform base model representations at nearly all layers across both datasets~\ref{fig:layerGraph}. Peak performance occurs in later layers for document-tuned transformers (Layer 21 on LMHD; Layer 19 on DS4UD) and earlier-to-mid layers for base models (Layer 19 on LMHD; Layer 10 on DS4UD). This pattern is consistent across datasets, suggesting document-level tuning objectives may preserve semantically relevant information in higher layers.

We compare models at their best-performing layers. Document-level representations achieve higher overall predictive performance on both datasets ($\Delta r = .012$ on LMHD; $\Delta r = .055$ on DS4UD; Table~\ref{tab:sent_token_delta}). Larger effect sizes in the DS4UD dataset could be due to the additional information of many more documents per person than in the LMHD dataset (DS4UD messages per person $\approx 85$, LMHD messages per person $\approx 6$)

\begin{figure}[t!]
     \centering
     \includegraphics[width=.5\textwidth]{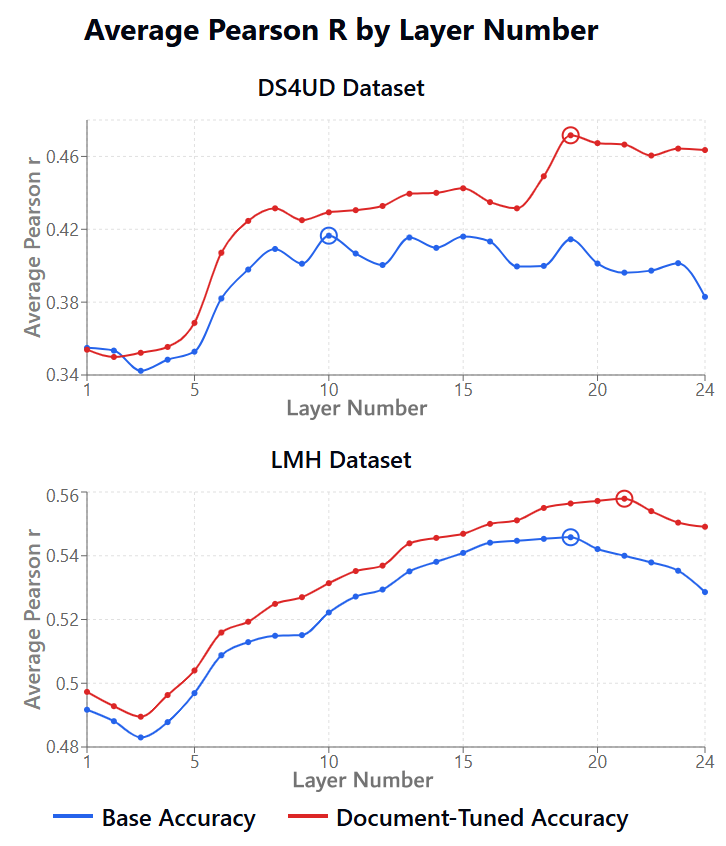}
     \caption{Prediction performance on the DS4UD (\textit{N=120}) and LMHD (\textit{N=1307}) datasets from different layer representations within "roberta-large" (Base Accuracy) and "all-roberta-large-v1" (Document-Tuned Accuracy)}
     \label{fig:layerGraph}
\end{figure}

\subsection{Layer-wise Performance Analysis}

Table~\ref{tab:topLayers} reports the top-performing layers (by Pearson's $r$) for both"roberta-large" (base) and "all-roberta-large-v1" (document-tuned) across datasets. This suggests that contrastive fine-tuning reorganizes semantic information toward the final layers, whereas MLM pretraining distributes it more evenly across depth — consistent with the view that later layers in document-tuned transformers are more specialized for semantic similarity.

\begin{table}[t]
\centering
\small
\setlength{\tabcolsep}{2pt}
\renewcommand{\arraystretch}{1.05}

\begin{tabular}{
r
@{\hspace{2pt}\vrule\hspace{3pt}}
r@{\hspace{10pt}} d{4}@{\hspace{10pt}\vrule\hspace{3pt}}
r@{\hspace{10pt}} d{4}@{\hspace{10pt}\vrule\hspace{3pt}}
r@{\hspace{10pt}} d{4}@{\hspace{10pt}\vrule\hspace{3pt}}
r@{\hspace{10pt}} d{4}
}
\toprule
\multicolumn{9}{c}{\textbf{Optimal Layer and Overall Performance}} \\
\midrule
& \multicolumn{4}{c}{\textbf{DS4UD} (\textit{N=120})} & \multicolumn{4}{c}{\textbf{LMHD} (\textit{N=1,307})} \\
\cmidrule(lr){2-5}\cmidrule(lr){6-9}
\textbf{R}
& \textbf{L} & \multicolumn{1}{c}{\textbf{Base}}
& \textbf{L} & \multicolumn{1}{c}{\textbf{Doc}}
& \textbf{L} & \multicolumn{1}{c}{\textbf{Base}}
& \textbf{L} & \multicolumn{1}{c}{\textbf{Doc}} \\
\midrule

1  & 10 & .4165       & 19 & .4716       & 19 & .5458       & 21 & .5579 \\
2  & 15 & .4160       & 20 & .4673       & 18 & .5453\sig   & 20 & .5572 \\
3  & 13 & .4155       & 21 & .4665       & 17 & .5447       & 19 & .5564\sig \\
4  & 19 & .4145       & 23 & .4643\sig   & 16 & .5441\sig   & 18 & .5550\sig \\
5  & 16 & .4133       & 24 & .4635\sig   & 20 & .5421\sig   & 22 & .5540\sig \\
6  & 14 & .4098       & 22 & .4605\sig   & 15 & .5409\sig   & 17 & .5511 \\
7  &  8 & .4092       & 18 & .4491\sig   & 21 & .5400\sig   & 23 & .5504\sig \\
8  & 11 & .4067       & 15 & .4425\sig   & 14 & .5381\sig   & 16 & .5500 \\
9  & 23 & .4014       & 14 & .4400\sig   & 22 & .5379\sig   & 24 & .5491\sig \\
10 & 20 & .4012       & 13 & .4395\sig   & 23 & .5353\sig   & 15 & .5469\sig \\

\bottomrule
\end{tabular}

\caption{Top-performing (Pearson $r$) layers for the DS4UD and LMHD dataset for "roberta-large" (Base) and "all-roberta-large-v1" (Doc). $\downarrow$ indicates significant distinction from the best performing layer ($p<.05$).}
\label{tab:topLayers}
\end{table}

\begin{table}[t]
\centering
\small
\setlength{\tabcolsep}{6pt}
\renewcommand{\arraystretch}{1.15}

\begin{tabular}{
l l
S[table-format=1.3, round-mode=places, round-precision=3]
S[table-format=1.3, round-mode=places, round-precision=3]
d{3}
}
\toprule
\multicolumn{5}{c}{\textbf{Model Performance}} \\
\midrule
\textbf{Outcome Category} & \textbf{Dataset} & \textbf{Doc} & \textbf{Base} & \multicolumn{1}{c}{\textbf{$\Delta$}} \\
\midrule
\multirow{2}{*}{\textbf{Overall}}
  & LMHD  & .572 & .561 & .012\textsuperscript{*} \\
  & DS4UD & .472 & .416 & .055\textsuperscript{*} \\
\bottomrule

\end{tabular}
\caption{Document-tuned vs.\ base model performance (Pearson $r$) by dataset.
$\Delta =$ document-tuned $-$ base. $^{*}p<.05$, $^{**}p<.01$.}
\label{tab:sent_token_delta}
\end{table}

\subsection{Model Performance by Domain} 
Improvements are not uniform across outcome domains (Table~\ref{tab:sent_token_delta}). Gains are most pronounced for affect, personality, and especially drinking behavior measures ($\Delta r = .081$ on LMHD and $.096$ on DS4UD), and more modest or inconsistent gains for mental health and demographic outcomes. The only outcome where base models outperformed document-tuned models was for predicting a users craving, otherwise, differences were small and non-significant, consistent with chance variation around a null effect.

Concatenating all user messages prior to embedding only slightly increases the advantage of document-tuned transformer embeddings, but relative to mean pooling, results are inconsistent, and on the LMHD dataset, drops both embedding types' predictive accuracy (Appendix table~\ref{tab:sent_token_msgcomb}).

\begin{table}[t]
\centering
\small
\setlength{\tabcolsep}{6pt}
\renewcommand{\arraystretch}{1.15}

\begin{tabular}{
l l
S[table-format=1.3, round-mode=places, round-precision=3]
S[table-format=1.3, round-mode=places, round-precision=3]
d{4}
}
\toprule
\multicolumn{5}{c}{\textbf{Model Performance by Message Combination Method}} \\
\midrule
\textbf{Dataset} & \textbf{Message Comb.}
& \textbf{Doc} & \textbf{Base} & \multicolumn{1}{c}{\textbf{$\Delta$}} \\
\midrule

\multirow{2}{*}{LMHD}
  & Mean Pool    & .572 & .561 & .012\textsuperscript{*} \\
  & Concatenate & .546 & .532 & .014\textsuperscript{*} \\
\midrule

\multirow{2}{*}{DS4UD}
  & Mean Pool    & .472 & .416 & .055\textsuperscript{*} \\
  & Concatenate & .482 & .411 & .071\textsuperscript{**} \\
\bottomrule

\end{tabular}

\caption{Document-tuned vs.\ base performance (Pearson $r$) by dataset and message combination.
$\Delta =$ document-tuned $-$ base. $^{*}p<.05$, $^{**}p<.01$.}
\label{tab:sent_token_msgcomb}

\end{table}

\subsection{Supervised Dimension Projection}
To further contextualize these findings, we qualitatively examine Supervised Dimension Projection Plots of specific words. These plots are constructed by computing an embedding vector representing the difference between high and low values of a given outcome. Individual words are then projected onto this axis using dot product projections, indicating how strongly each word is associated with the outcome in a given embedding space \cite{kjell2023text}. To compare two models' embeddings of the same dataset, we plot the word along an embedding dimension representing the outcome in both embedding spaces, resulting in a 2D graph. We find no consistent trends in the number of significantly predictive words for either model. While graphs show both expected behavior (like "happy" being equally low in both representations), it also shows words like "Alright" in the second quadrant, which maps to high PHQ scores in document-tuned embeddings and low in base embeddings. This suggests that document-tuned models may encode the word 'Alright' in a context more associated with 'not being alright', whereas base models may place more emphasis independently on its definition alone — potentially reflecting the richer contextual representations learned through contrastive training. Embedded hedged language (e.g., `usually' in PHQ and 'pretty' in AUDIT) tended to be more predictive of outcomes in document-tuned embeddings while abundance (e.g., `lot') was more predictive in traditional word-level models, suggesting document-tuned models may better capture uncertainty and tentativeness in self-report language.

\begin{figure*}[t!]
     \centering
     \includegraphics[width=1\textwidth]{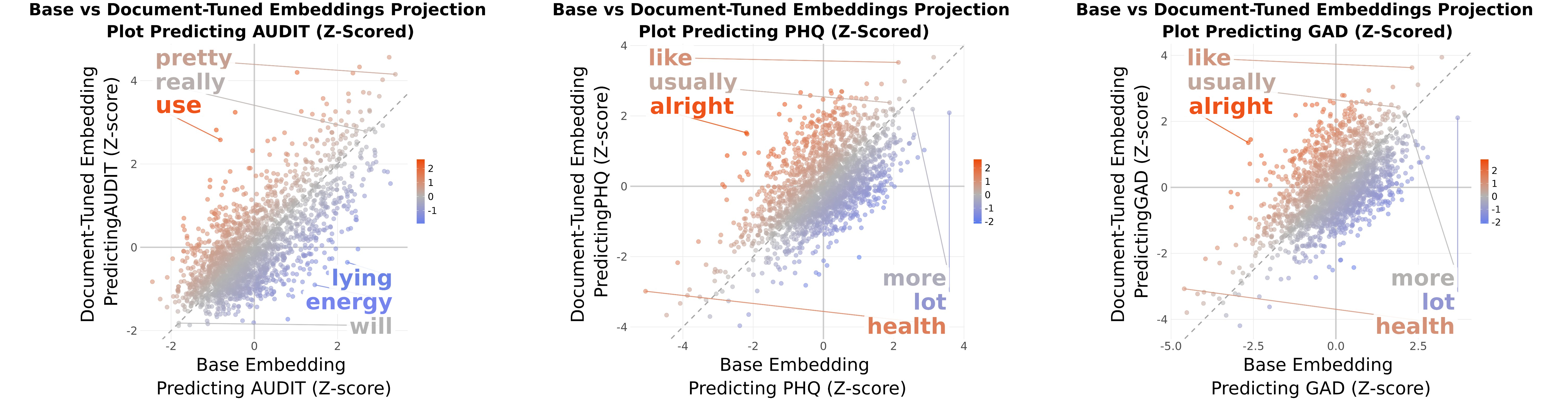}
     \caption{Dimension Projection Plot where words from the LMHD dataset are plotted along the embedding vector representing a difference in High vs Low PHQ. Words plotted along the diagonal line are represented similarly in both embedding spaces with relation to PHQ. More saturated points (like "meaning") are associated with high PHQ in one space and low PHQ in the other. Orange points are more highly associated with the outcome in Document Tuned embeddings, and blue points are more highly associated with the outcome in Base embeddings.}
     \label{fig:layerGraph}
\end{figure*}

\subsection{Perturbations}
In recent work, such as \cite{alahmari2025large}, models have been probed for their robustness to human-induced linguistic errors that are common in real-world applications. As natural language surveys and EMAs are prone to user error, we corrupt text using deletion, typo injection, synonym replacement, and back translation at varying levels to simulate increasingly noisy data. We find in most cases, neither document-tuned nor base transformers are more robust to these perturbations than the other. Across deletion, typo injection, and back translation, both models showed comparable and gradual performance degradation as corruption levels increased. Synonym replacement shows slightly more robustness by the base transformer at extremely high percentages of replaced words, potentially because base model embeddings are less sensitive to global semantic context, making synonym substitution less disruptive to the overall representation. This fails to explain the stronger overall performance of document-tuned models, suggesting that their advantage stems from a more fundamental difference in how meaning is encoded that does not simply make the model more robust to syntactic or semantic noise.

\begin{figure}[t!]
     \centering
     \includegraphics[width=.48 \textwidth]{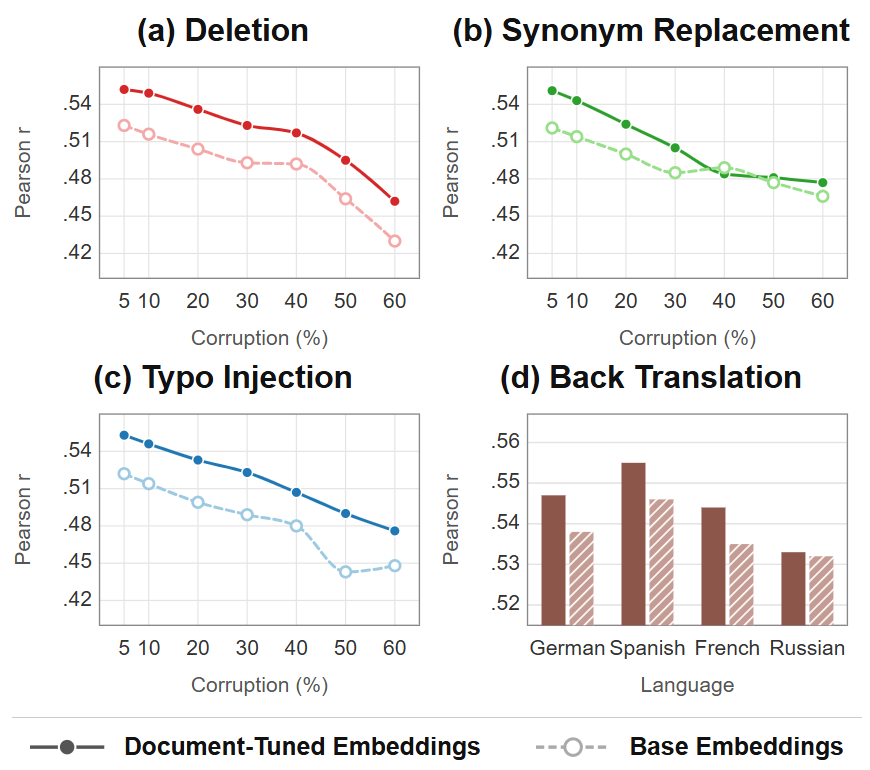}
     \caption{Varying Levels of Text Perturbations vs Model Prediction Correlation with True Scores. (a) consists of deleting varying percentages of words. (b) consists of adding, replacing, or reordering adjacent letters in varying percentages of words. (c) consists of using WordNet to replace varying percentages of words with synonyms. (d) consists of translating the message into another language and back again.}
     \label{fig:perturbations}
\end{figure}

\section{Conclusion}

This work presents a controlled comparison of base transformers (utilizing token representations) and document-tuned transformers (utilizing document-level, contrastively tuned representations) for psychological prediction. 
Document-tuned transformers consistently outperform base transformers for user-level psychological prediction across two longitudinal datasets. Gains are modest, but statistically significant and robust across layers. 
Optimal performance occurs in later layers for document-tuned models, with the final four to five layers performing comparably. We examined several candidate explanations for the observed performance differences, including outcome type, word embedding representation, and robustness to linguistic perturbations, but found no consistent moderating patterns. 
Crucially, the advantage of document-tuned representations was not confined to any particular outcome domain, suggesting a broadly superior capacity to encode person-level meaning rather than sensitivity to specific linguistic features — making document-tuned models a reasonable default choice for language-based mental health assessment. For researchers developing language-based screening or monitoring systems -- particularly those using EMA or open-ended survey responses -- these results provide empirical guidance on a design decision that is typically made without systematic comparison: which encoder to use and which layers to extract embeddings from.

\section{Ethical Considerations}
Both datasets were collected under institutional review board (IRB) approval with informed consent from participants. Analyses were conducted on de-identified text and survey data. 

In particular, the improved prediction of sensitive outcomes such as drinking behavior and suicidality from open-ended text raises questions about the potential for unintended inference in deployed systems. Models developed in this work should not be used for clinical diagnosis or high-stakes decision-making without appropriate validation and oversight. While our study focuses on methodological comparison rather than deployment, representation choices that improve predictive performance may also increase risks of unintended inference. Care should be taken to ensure that such systems are used transparently, with appropriate safeguards, and in ways that respect participant privacy and autonomy.

\section{Limitations}

Several limitations should be noted. First, our comparison uses off-the-shelf checkpoints. "All-roberta-large-v1" includes additional post-pretraining beyond masked language modeling, involving different data. As such, observed gains cannot be attributed solely to the contrastive objective without compute- and data-matched controls.

Second, we evaluate only two English-language longitudinal datasets, including one relatively small sample (N=120). Results may not generalize to other populations, languages, platforms (e.g., clinical notes or social media), or measurement settings.

Third, our evaluation focuses on predictive accuracy (Pearson’s $r$) and does not assess calibration, fairness, or error disparities across demographic groups, which are important considerations for assessment contexts.

Fourth, though this dataset contains longitudinal data, we chose to measure cross sectional prediction accuracy for simplicity and ease of comparison between the datasets. We leave studying longitudinal models of psychological constructs to future work.

Finally, this study evaluates methodological differences rather than real-world deployment. Improvements in predictive performance do not imply clinical validity or suitability for high-stakes decision-making.

\bibliography{custom}

\appendix

\section{Appendix}

\subsection{Full Data Sizes}

Table~\ref{tab:dataset_stats_full} summarizes data from both datasets.

\begin{table}[h!]
\centering
\small
\setlength{\tabcolsep}{6pt}
\renewcommand{\arraystretch}{1.15}

\begin{tabular}{llrr}
\toprule
& & \multicolumn{2}{c}{\textbf{EMAs}} \\
\cmidrule(lr){3-4}
\textbf{Dataset} & \textbf{Statistic}
& \textbf{Indiv.}
& \textbf{Concat.} \\
\midrule

\multirow{4}{*}{DS4UD}
& Avg. EMA per emb.   & 1       & 6.9 \\
& Avg. emb. per user   & 84.9       & 12.4 \\
& Total embeddings               & 10{,}108 & 1{,}486 \\
& Total users              & 120      & 120 \\

\midrule

\multirow{4}{*}{LMHD}
& Avg. EMA per emb.   & 1       & 5.2 \\
& Avg. emb. per user   & 5.5       & 1.1 \\
& Total embeddings           & 7{,}207  & 1{,}391 \\
& Total users              & 1{,}307  & 1{,}307 \\

\bottomrule

\end{tabular}
\caption{Descriptive statistics for the DS4UD and LMHD datasets under individual-message and concatenated-message representations.}
\label{tab:dataset_stats_full}
\end{table}

\subsection{Outcome-Level Results by Domain}

Table~\ref{tab:sent_token_delta_full} summarizes performance by outcome category.

\begin{table}[h!]
\small
\centering
\setlength{\tabcolsep}{6pt}
\renewcommand{\arraystretch}{1.15}

\begin{tabular}{
l l
S[table-format=1.3, round-mode=places, round-precision=3]
S[table-format=1.3, round-mode=places, round-precision=3]
d{3}
}
\toprule
\multicolumn{5}{c}{\textbf{Model Performance by Outcome Category}} \\
\midrule
\textbf{Outcome Category} & \textbf{Dataset} & \textbf{Doc} & \textbf{Base} & \multicolumn{1}{c}{\textbf{$\Delta$}} \\
\midrule

\rowcolor{gray!12}
Affect
  & DS4UD & .502 & .432 & .070\textsuperscript{*} \\

Demographic
  & DS4UD & .489 & .429 & .061 \\

\rowcolor{gray!12}
Drinking behavior
  & LMHD  & .263 & .182 & .081\textsuperscript{*} \\
\rowcolor{gray!8}
Drinking behavior
  & DS4UD & .337 & .241 & .096\textsuperscript{**} \\

Mental health
  & LMHD  & .576 & .566 & .010\textsuperscript{*} \\
Mental health
  & DS4UD & .636 & .615 & .021 \\

\rowcolor{gray!12}
Personality
  & DS4UD & .378 & .337 & .041\textsuperscript{*} \\

\midrule
\multirow{2}{*}{\textbf{Overall}}
  & LMHD  & .572 & .561 & .012\textsuperscript{*} \\
  & DS4UD & .472 & .416 & .055\textsuperscript{*} \\

\bottomrule

\end{tabular}
\caption{Document-Tuned vs.\ Base model performance (Pearson $r$) by outcome category and dataset.
$\Delta =$ Document-Tuned $-$ Base $^{*}p<.05$, $^{**}p<.01$.}
\label{tab:sent_token_delta_full}
\end{table}

\subsection{Full Outcome-Level Results}

Table~\ref{tab:full_results} reports prediction performance for all individual outcomes across both datasets.

\begin{table*}[t]
\centering
\scriptsize
\setlength{\tabcolsep}{5pt}
\renewcommand{\arraystretch}{1.15}
\begin{tabular}{
l r
S[table-format=2.2, parse-numbers=false]
S[table-format=2.2, parse-numbers=false]
d{3}
S[table-format=.3, parse-numbers=false]
S[table-format=.3, parse-numbers=false]
d{3}
}
\toprule
\multicolumn{8}{c}{\textbf{Model Performance}} \\
\midrule
\textbf{Outcome} & \textbf{N}
  & \multicolumn{3}{c}{\textbf{MAE}}
  & \multicolumn{3}{c}{\textbf{Pearson $r$}} \\
\cmidrule(lr){3-5} \cmidrule(lr){6-8}
  & & \textbf{Doc} & \textbf{Base} & \multicolumn{1}{c}{\textbf{$\Delta$}}
    & \textbf{Doc} & \textbf{Base} & \multicolumn{1}{c}{\textbf{$\Delta$}} \\
\midrule
\multicolumn{8}{l}{\textbf{LMHD}} \\
\midrule
AUDIT                  & 1307 & 3.63  & 3.70  & -.07\textsuperscript{*}  & .263 & .182 & .081\textsuperscript{*}  \\
Age                    & 1307 & 8.62  & 8.14  & .48                      & .792 & .812 & -.020                   \\
DUDIT                  & 1307 & 1.79  & 1.78  & .01                      & .181 & .156 & .025                    \\
GAD2                   & 1307 & 1.12  & 1.10  & .02                      & .662 & .673 & -.011                   \\
GAD                    & 1307 & 3.05  & 3.09  & -.04                     & .724 & .725 & -.001                   \\
HILS                   & 1307 & 2.87  & 2.88  & -.01                     & .710 & .705 & .005                    \\
HCMental       &  708 & 0.83  & 0.83  & .00                      & .339 & .349 & -.011                   \\
HCVisits       & 1307 & 1.34  & 1.33  & .01                      & .357 & .321 & .036 \\
IDAS\_AppGain     & 1307 & 2.63  & 2.64  & -.01                     & .300 & .291 & .009                    \\
IDAS\_AppLoss     & 1307 & 2.08  & 2.08  & .00                      & .412 & .415 & -.004                   \\
IDAS\_Dysphoria        & 1307 & 5.05  & 5.04  & .01                      & .752 & .753 & -.001                   \\
IDAS\_Temper        & 1307 & 2.41  & 2.47  & -.06\textsuperscript{*}  & .573 & .542 & .031\textsuperscript{*} \\
IDAS\_Insomnia         & 1307 & 4.55  & 4.59  & -.04                     & .475 & .459 & .016                    \\
IDAS\_Lassitude        & 1307 & 3.14  & 3.09  & .05                      & .676 & .686 & -.009                   \\
IDAS\_Panic            & 1307 & 3.24  & 3.35  & -.11\textsuperscript{*}  & .585 & .562 & .022\textsuperscript{*} \\
IDAS\_SocAnx   & 1307 & 3.15  & 3.13  & .02                      & .610 & .614 & -.005                   \\
IDAS\_Suic      & 1307 & 1.94  & 1.97  & -.03 & .604 & .576 & .028 \\
IDAS\_Traum     & 1307 & 2.31  & 2.36  & -.05\textsuperscript{*}  & .576 & .545 & .031\textsuperscript{*} \\
IDAS\_WB        & 1307 & 4.87  & 5.01  & -.14\textsuperscript{*}  & .670 & .665 & .005\textsuperscript{*}                    \\
MentalSickDays         &  395 & 12.56 & 12.64 & -.08                     & .390 & .354 & .036 \\
PHQ2\_sum              & 1307 & 0.89  & 0.90  & -.01                     & .712 & .705 & .007                    \\
PHQtot                 & 1307 & 3.29  & 3.32  & -.03                     & .744 & .739 & .005                    \\
PSS4\_sum              & 1263 & 2.11  & 2.11  & .00                      & .692 & .688 & .004                    \\
PSStot                 & 1307 & 4.87  & 4.89  & -.02                     & .741 & .740 & .001                    \\
SWLStot                & 1307 & 3.10  & 3.14  & -.04                     & .683 & .676 & .007                    \\
SickDaysMonth          & 1307 & 5.63  & 5.56  & .07\textsuperscript{*}   & .206 & .189 & .018\textsuperscript{*} \\
WHODAS       & 1307 & 17.47 & 17.96 & -.49\textsuperscript{*}  & .636 & .615 & .021\textsuperscript{*} \\
\midrule
\textit{Mean} &      & 4.02  & 4.04  & -.02\textsuperscript{*}                     & .572 & .561 & .012\textsuperscript{*} \\
\midrule
\multicolumn{8}{l}{\textbf{DS4UD}} \\
\midrule
affect                      & 120 & 0.35  & 0.39  & -.04\textsuperscript{*}  & .799 & .752 & .047\textsuperscript{*} \\
age                         & 103 & 5.71  & 5.95  & -.24  & .500 & .425 & .074 \\
agreeable             & 120 & 1.75  & 1.76  & -.01                     & .455 & .393 & .063 \\
anxiety               & 120 & 2.65  & 2.70  & -.05                     & .582 & .568 & .014                    \\
audit10                 & 103 & 5.41  & 5.70  & -.29\textsuperscript{*}  & .318 & .130 & .188\textsuperscript{*} \\
auditc                  & 120 & 1.94  & 2.07  & -.13\textsuperscript{*}  & .310 & .214 & .097\textsuperscript{*} \\
conscientious         & 120 & 2.01  & 2.10  & -.09\textsuperscript{*}                     & .392 & .372 & .020\textsuperscript{*}                    \\
craving 1                   &  94 & 1.21  & 1.19  & .02                      & .403 & .439 & -.036 \\
depression score            & 120 & 3.90  & 3.81  & .09   & .600 & .601 & -.002                   \\
energy                      & 120 & 0.24  & 0.25  & -.01\textsuperscript{*}                    & .283 & .129 & .154\textsuperscript{*} \\
extravert             & 120 & 2.64  & 2.63  & .01                      & .320 & .304 & .017                    \\
gad7 sum                    & 120 & 3.64  & 3.65  & -.01                     & .610 & .567 & .042 \\
income           & 120 & 2.00  & 2.01  & -.01                     & .479 & .432 & .048 \\
mace                    &  99 & 9.26  & 10.27 & -1.01\textsuperscript{*} & .339 & .160 & .179\textsuperscript{*} \\
neg affect            & 103 & 3.03  & 3.12  & -.09 & .536 & .510 & .026 \\
neurotic              & 120 & 2.00  & 2.10  & -.10& .632 & .586 & .046 \\
openness              & 120 & 2.23  & 2.23  & .00                      & .089 & .030 & .059 \\
phq9                    & 120 & 3.72  & 3.77  & -.05                     & .601 & .594 & .007                    \\
pos affect            & 103 & 3.21  & 3.38  & -.17\textsuperscript{*}  & .389 & .335 & .054\textsuperscript{*} \\
pss nerv. stress agr.       &  92 & 0.57  & 0.62  & -.05\textsuperscript{*}  & .784 & .722 & .061\textsuperscript{*} \\
pss                   & 120 & 1.66  & 1.67  & -.01                     & .639 & .635 & .004                    \\
unhealthy drinking          & 120 & 1.28  & 1.31  & -.03  & .314 & .264 & .050 \\
\midrule
\textit{Mean}      &     & 2.75  & 2.85  & -.10\textsuperscript{*}  & .472 & .416 & .055\textsuperscript{*} \\
\bottomrule
\end{tabular}
\vspace{-7pt}
\caption{Document-tuned vs.\ base performance by outcome and dataset.
MAE = Mean Absolute Error (lower is better); Pearson $r$ (higher is better).
$\Delta =$ Document-tuned $-$ Base. $^{*}p<.05$, $^{**}p<.01$ (significances from t-test with MAE).}
\label{tab:full_results}
\end{table*}

\end{document}